\documentclass{article}
\usepackage[preprint]{neurips_2026}

\usepackage[utf8]{inputenc}
\usepackage[T1]{fontenc}
\usepackage{hyperref}
\usepackage{url}
\usepackage{booktabs}
\usepackage{amsfonts}
\usepackage{nicefrac}
\usepackage{microtype}
\usepackage{xcolor}
\usepackage{graphicx}
\usepackage{amsmath}
\usepackage{amssymb}
\usepackage{placeins}
\usepackage{wrapfig}
\usepackage{enumitem} 
\definecolor{tokenblue}{HTML}{DDE7FA}
\definecolor{timepink}{HTML}{F0C9CC}
\definecolor{iteryellow}{HTML}{FCF2C4}

\usepackage{tikz}
\usetikzlibrary{arrows.meta,positioning}

\newcommand{\name}{SkillSmith}

\title{\name{}: Compiling Agent Skills into Boundary-Guided Runtime Interfaces}

\author{
\begin{tabular}{c}
\textbf{Duling Xu}$^{1,2}$ \quad
\textbf{Zheng Chen}$^{1}$\thanks{Corresponding author} \quad
\textbf{Zaifeng Pan}$^{3}$ \quad
\textbf{Jiawei Guan}$^{2}$ \\
\textbf{Dong Dong}$^{1}$ \quad
\textbf{Jialin Li}$^{1}$ \quad
\textbf{Bangzheng Pu}$^{1}$ \\
\textsuperscript{1} AetherHeart Tech Co., Ltd., Shanghai, China \\
\textsuperscript{2} Renmin University of China, Beijing, China \\
\textsuperscript{3} University of California San Diego, CA, USA \\
\texttt{\{duling, zchen, dongd, lijialin, bzpu\}@aetherheart.com} \\
\texttt{zapan@ucsd.edu} \quad
\texttt{guanjw@ruc.edu.cn}
\end{tabular}
}

\begin{document}

\maketitle

\begin{abstract}
Recently, \emph{skills} have been widely adopted in large language model (LLM)-based agent systems across various domains. In existing frameworks, skills are typically injected into the agent reasoning loop as contextual guidance once matched to a runtime task, enabling specialized task-solving capabilities. 
We find that this execution paradigm introduces two major sources of redundancy: irrelevant context injection and repeated skill-specific reasoning and planning.
To this end, we propose \name{}, a boundary-first compiler-runtime framework that compiles skill packages offline into minimal executable interfaces. 
By extracting fine-grained operational boundaries from skills, \name{} enables agents to dynamically access and execute only the relevant components at runtime, thereby minimizing unnecessary context injection and redundant reasoning overhead.
In the evaluation on SkillsBench benchmark, \name{} reduces
solve-stage token usage by 57.44\%, thinking iterations by 42.99\%, solve time
by 50.57\% (2.02$\times$ faster), and token-proportional monetary cost by
57.44\% compared with using raw-skills.
Moreover, compiled artifacts produced by a stronger model can be reused by a
smaller or more efficient runtime model, improving task accuracy in cases where
raw skill interpretation fails.
The source code and data are available at \url{https://github.com/AetherHeart-AI/Aeloon}.
\end{abstract}

\section{Introduction}
Large language model (LLM)-based AI agents have recently attracted widespread attention for their ability to solve real-world tasks through reasoning, planning, tool use, and iterative execution~\cite{yao2022react, wei2022chain, yao2023tree, schick2023toolformer, qin2023toolllm, madaan2023self, shinn2023reflexion}. Such systems have demonstrated strong performance across domains including coding, mathematical problem solving, and daily office workflows~\cite{yao2022webshop, wang2023voyager}.
Recently, the concept of \emph{skills} has rapidly gained popularity in the agent community due to its strong performance and portability~\cite{NEURIPS2024_e4c61f57, li2026skillsbench}. A skill represents developer-authored expertise for handling a particular class of tasks and is typically distributed as a reusable package containing a \texttt{SKILL.md} instruction file together with auxiliary resources such as scripts, templates, configuration files, examples, and supporting references.

\begin{figure}[t]
\centering
\includegraphics[width=1\linewidth]{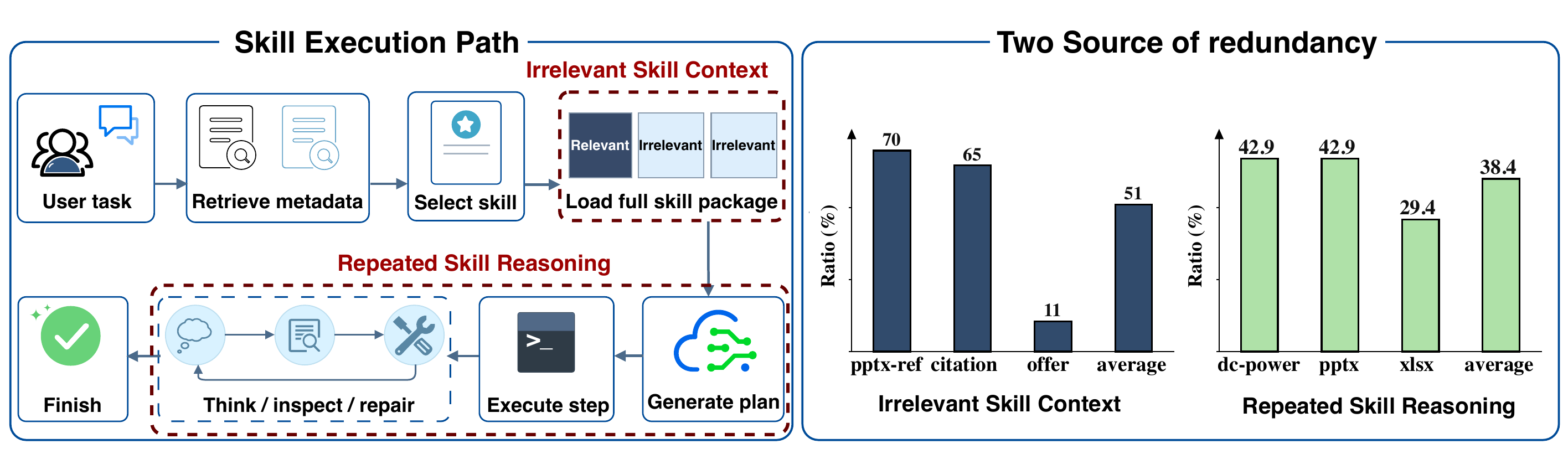}
\caption{Skills-enabled execution path and its two source of redundancy.}
\label{fig:skill_runtime_cost}
\end{figure}

In contemporary agent frameworks, skills are typically integrated as instructional context within the ReAct-style~\cite{yao2022react} reasoning loop, as shown in Figure~\ref{fig:skill_runtime_cost}.
When the model determines that the current runtime task matches a particular skill, the agent retrieves and injects the full skill information into the model context. 
The model then reasons over the injected material to plan the next execution steps, selects and invokes appropriate actions or tools, observes the resulting feedback, and repeats this reasoning–action cycle iteratively until task completion.

Despite their effectiveness for specialized tasks, large skill packages also introduce substantial runtime redundancy. 
Figure~\ref{fig:skill_runtime_cost} summarizes and illustrates two sources of redundancy.
The first source is \emph{irrelevant skill context}. 
In existing agent systems, once a skill is selected, the entire skill package—including \texttt{SKILL.md} files and bundled assets—is typically injected into the model context before execution. 
However, only a subset of this information is actually relevant to the current runtime task. 
Across seven SkillsBench~\cite{li2026skillsbench} tasks, agents load approximately 17.8K source tokens per execution, of which 9.1K tokens (51.21\%) are ultimately irrelevant to the observed execution process.
This suggests that a substantial portion of runtime context consumption comes from unnecessary skill content.
The second source is \emph{repeated skill reasoning}. 
After loading a skill, the model repeatedly interprets the skill instructions and reconstructs an execution strategy online. We find that different tasks using the same skill exhibit an average reasoning-trace similarity of 45.5\%. 
This similarity suggests that, for the same skill, the model repeatedly spends computation on understanding the skill and regenerating highly similar execution plans across different runtime tasks.


We argue that both sources of redundancy stem from the lack of explicit runtime boundaries in existing skill systems. 
Today, skills are treated as monolithic textual resources that must be repeatedly interpreted online, causing agents to reload irrelevant context and reconstruct similar execution structures across tasks.

Inspired by the separation of offline transformation and runtime execution in traditional compiler systems, we explore whether parts of this interpretation process can be shifted offline. 
However, unlike conventional programs, natural-language skill packages are highly heterogeneous in structure, semantics, and operational form. 
Some skills primarily act as descriptive instruction manuals, while others resemble executable workflows, procedural checklists, or tool-operation guides. 
As a result, skills cannot be reliably lowered into a single standardized intermediate representation with fixed frontend-to-backend compilation passes.

To address this challenge, we propose \textbf{\name{}}, a boundary-first compiler-runtime framework for agent skills. 
Rather than compiling skills into a unified workflow IR, \name{} compiles them into explicit runtime boundary contracts: minimal executable interfaces that capture the operationally useful portions of a skill. 
The compiler analyzes skill-local capability structure, extracts locally compilable fragments, and transforms them into normalized runtime artifacts exposing executable operators, input requirements, policy constraints, validation evidence, and fallback paths. 
These boundary contracts become the runtime representation consumed by the agent, enabling selective disclosure and execution while avoiding repeated interpretation of the original skill source.

At runtime, \name{} uses this boundary contract through progressive disclosure.
The agent first sees a compact skill handle and boundary summary; detailed
operators, policies, and fallback contents are disclosed only after the agent
selects the compiled skill.  The shared runtime then selects the relevant
operator, checks policy constraints, executes typed operations when safe,
returns guidance when direct execution is inappropriate, and falls back to
preserved source material when further reasoning is required.

Across representative agent harnesses and runtime models, on average over seven SkillsBench tasks, 
\name{} reduces solve-stage token
usage by 57.44\%, reasoning/LLM calls by 42.99\%, solve time by 50.57\%
(2.02$\times$ faster), and token-proportional monetary cost by 57.44\%
compared with Raw-Skills.  
Against SkVM~\cite{chen2026skvm}, a recent skill virtual machine that
compiles skills for heterogeneous LLMs and agent harnesses, \name{} reduces
tokens by 46.49\%, solve time by 47.04\%, and calls by 18.67\%.
In terms of task accuracy, \name{} does not introduce observable correctness regressions on successful Raw-Skills cases, and further improves task success on medium-scale runtime models through artifacts compiled with a stronger model.

Our contributions are as follows:
\begin{itemize}[leftmargin=2.2em]

    \item We identify two major sources of runtime redundancy introduced by skill-based agent systems: irrelevant skill context injection and repeated reasoning over similar execution plans.
    \item We propose \name{}, a boundary-first compiler-runtime framework that compiles skill packages into minimal executable runtime units.
    \item We demonstrate on a difficulty-stratified set of representative SkillsBench tasks that SkillSmith significantly reduces token consumption, reasoning latency, and the number of model reasoning iterations, while maintaining or even improving task accuracy.

\end{itemize}

\section{Related Work}

\textbf{Agent skills and procedural knowledge.}
Recent agent systems increasingly package domain procedures as reusable skills:
filesystem directories containing a \texttt{SKILL.md} file, scripts, templates,
and supporting references~\cite{anthropic2025agentskills,anthropic2026agentskillsdocs}.
This format gives agents on-demand access to specialized procedural knowledge
without requiring users to repeat the same instructions in every conversation.
SkillsBench~\cite{li2026skillsbench} further establishes skills as first-class
evaluation artifacts by pairing diverse tasks with curated skill packages and
deterministic verifiers. These systems demonstrate that skills are useful units
of reuse, but they primarily treat skills as contextual resources to be loaded
and interpreted by the agent at runtime. \name{} instead targets the execution
layer: it transforms skill specifications into structured workflow artifacts
that can be executed, inspected, and resumed across invocations.

\textbf{LLM agents and tool-oriented orchestration.}
LLM agents commonly rely on the model as the online controller. ReAct~\cite{yao2022react}
interleaves reasoning with tool actions, while AutoGPT~\cite{autogpt2023}
popularized autonomous task decomposition. Tool-use work such as
Toolformer~\cite{schick2023toolformer}, ToolLLM~\cite{qin2023toolllm}, and
Gorilla~\cite{NEURIPS2024_e4c61f57} improves API or tool selection. Agent
frameworks such as AutoGen~\cite{wu2023autogen} and Magentic-One~\cite{fourney2024magenticone}
compose multiple agents and tools to solve complex tasks. These approaches
provide flexible orchestration, but the concrete execution path is usually
planned and revised online by the agent. In contrast, \name{} moves reusable
skill understanding out of repeated runtime reasoning and into a reusable
workflow artifact.

\textbf{Constrained and program-aided execution.}
A related line of work reduces free-form model execution by delegating parts of
the computation to structured substrates. PAL~\cite{gao2023pal} uses generated
Python programs for reasoning tasks; LMQL~\cite{beurer2023prompting} constrains
language-model programs through query-level control flow; SayCan~\cite{ahn2022can}
grounds language-model choices in robotic affordances; and Voyager~\cite{wang2023voyager}
stores reusable programs for open-ended embodied tasks. These works share the
principle that not every part of a task should remain open-ended LLM reasoning.
However, they focus on program-aided reasoning, constrained generation, robotic
skill selection, or accumulated code libraries rather than compiling general
agent skill packages into resumable workflow runtimes.

\textbf{Compilation and runtime systems for LLM programs.}
Several systems bring programming-language and runtime ideas to LLM applications.
DSPy~\cite{khattab2023dspy} compiles declarative LM pipelines by optimizing
prompts and demonstrations for a target metric. SGLang~\cite{NEURIPS2024_724be447}
provides a frontend and runtime for efficient execution of structured language
model programs. Workflow runtimes such as LangGraph provide persistence and
checkpointing for manually specified agent graphs~\cite{langgraph2026persistence}.
Recent skill-runtime work such as SkVM~\cite{chen2026skvm} explores compiler-style
optimization for agent skills, emphasizing portability across heterogeneous
models and harnesses through capability profiling, skill rewriting, environment
binding, concurrency extraction, and runtime optimization. \name{} is
complementary to these efforts: rather than optimizing prompt pipelines,
inference runtimes, or cross-harness portability, it compiles skill
specifications into executable workflow artifacts for efficient and resumable
execution within an agent workspace.

\textbf{Positioning.}
Existing systems either treat skills as runtime context, evaluate whether skills
help agents, orchestrate online agent reasoning, optimize prompt or generation
pipelines, or provide workflow runtimes for developer-authored graphs. \name{}
focuses on a different layer: converting reusable skill specifications into
structured, source-grounded, and resumable workflow artifacts. This reduces
repeated skill interpretation while preserving selective LLM invocation for
steps that genuinely require generation.

\section{Method}

\begin{figure}
  \centering
  \includegraphics[width=\linewidth]{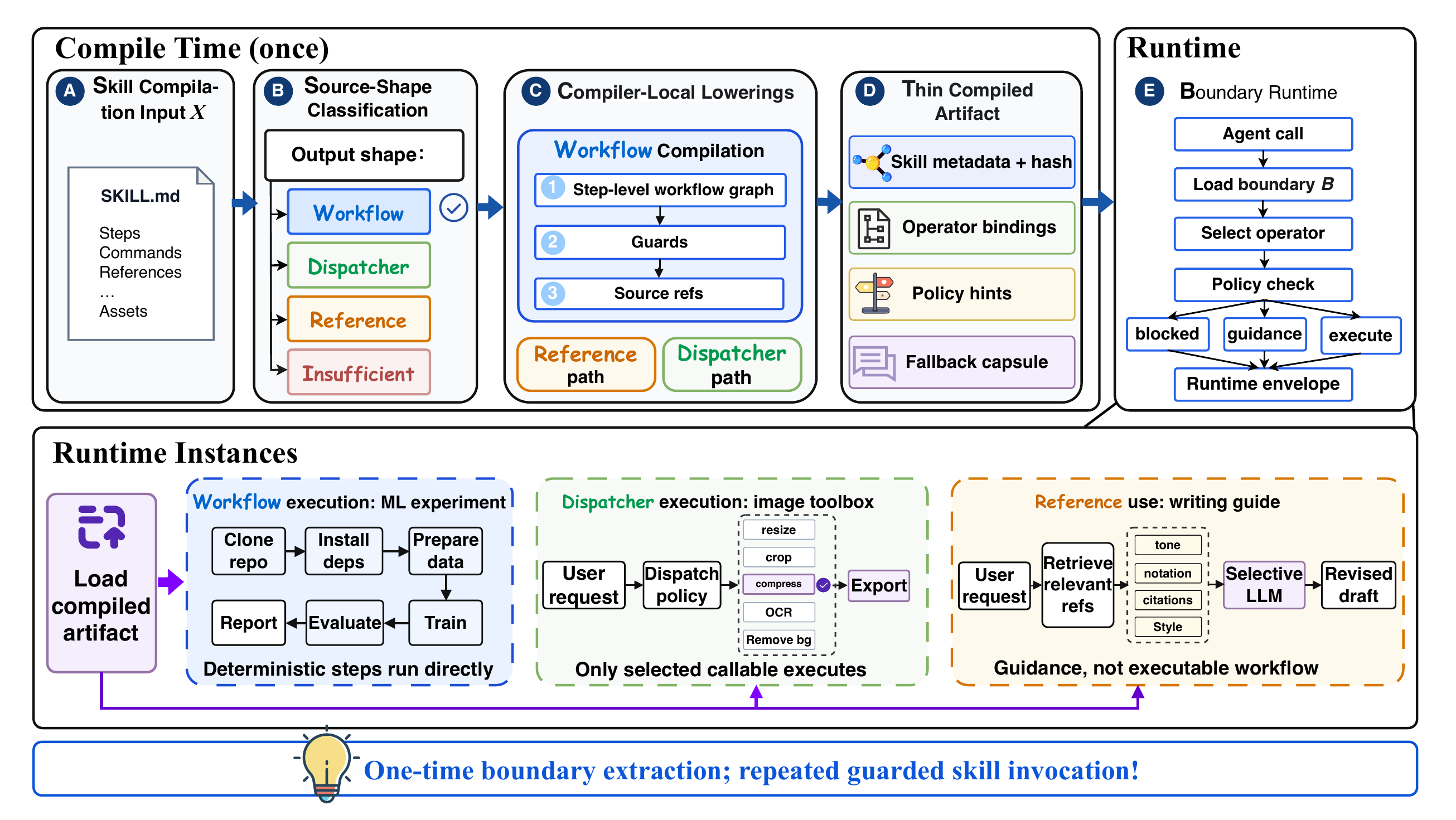}
  \vspace{-0.5cm}
  \caption{\name{} system overview.}
  \label{fig:method_overview}
\end{figure}

\name{} treats skills as compilable capability specifications rather than
prompt fragments.  
Its central design choice is \emph{boundary-first
compilation}: \name{} compiles each skill package into an explicit runtime
contract that captures the artifact's policy-governed contribution boundary, invocation interface, execution conditions, and fallback obligations.

The contract, rather than a workflow IR, is the public compilation target;
workflow graphs, dispatcher extractions, and reference indexes are internal
lowerings used to construct that interface.
This contract makes
skill execution explicit without requiring the compiled artifact to subsume the
entire task, allowing the agent to combine compiled skill behavior with normal
reasoning and fallback tools.  
Figure~\ref{fig:method_overview} places this
design in the full compile-time and runtime pipeline; the rest of this section
then develops the boundary-aware formulation in detail.

\vspace{-0.2cm}
\subsection{Skill Packages as Compilation Inputs}
The compilation input to \name{} is a skill package together with the runtime
context under which the compiled artifact will be used.  We denote an input as
\[
X = (P, \mathcal{T}, \mathcal{E}, \Pi),
\]
where \(P\) is the skill package, \(\mathcal{T}\) is the available tool
interface, \(\mathcal{E}\) describes the execution environment, and \(\Pi\)
specifies compilation policies such as sandboxing, cache reuse, and whether
task-bound adaptation is permitted.

A skill package is defined as
\[
P = (d, \mathcal{A}, m, h).
\]
Here \(d\) is the entry \texttt{SKILL.md} document,
\(\mathcal{A}=\{a_i\}_{i=1}^n\) is the set of package-local assets, \(m\) is
package metadata, and \(h\) is a content hash over the canonicalized package
contents.  Each asset is represented as
\[
a_i = (p_i, \tau_i, s_i, \chi_i),
\]
where \(p_i\) is its package-relative path, \(\tau_i\) is its inferred asset
type, \(s_i\) is its byte size, and \(\chi_i\) is its \texttt{sha256} digest.
The package hash is computed from the entry document and all asset records, and
is used as both a cache key and a reproducibility boundary.  A compiled artifact
is valid only for the package hash from which it was produced.

Thus, \name{} compiles the full package boundary rather than the entry document
alone.  We use \(\Pi\) to specify which inputs the compiler may inspect during
compilation.  In the generic setting, the compiler builds artifacts only from
\(P\), \(\mathcal{T}\), and \(\mathcal{E}\).  At runtime, the compiled artifact
exposes operators that the agent dynamically selects and instantiates with the
current task inputs.

\vspace{-0.1cm}
\subsection{Source-Shape Classification and Lowering}

\name{} classifies each package before artifact generation, using evidence from
the entry document and package assets to determine the package's capability
shape.  The classifier checks for ordered procedures, command blocks, scripts,
API signatures, setup requirements, reference density, and explicit verification
cues, then chooses the compiler-local lowering in
Table~\ref{tab:source_shape_lowering}.
The resulting target preserves the package's natural capability shape while
exposing a uniform boundary contract in the next stage.

\begin{table}[t]
\centering
\small
\caption{Source-shape classification and compiler-local lowerings.}
\label{tab:source_shape_lowering}
\begin{tabular}{p{0.11\linewidth} p{0.44\linewidth} p{0.36\linewidth}}
\toprule
\textbf{Shape} & \textbf{Classification evidence} & \textbf{Compiler-local lowering} \\
\midrule
\emph{workflow} &
Ordered steps, control-flow headings, command sequences, input/output mentions,
or verification instructions. &
Step-level workflow graph with dependencies, execution types, guards, and
source references. \\
\emph{dispatcher} &
Bundled scripts, API or function descriptions, reusable command snippets, or
multiple callable operations without a fixed order. &
Dispatcher capabilities and typed operators selected dynamically by the agent. \\
\emph{reference} &
Reference-heavy prose, tables, formulas, examples, templates, or domain
guidance with limited executable structure. &
Indexed reference sections for retrieval and grounding. \\
\emph{insufficient} &
Missing entry content, ambiguous capability boundaries, or too little reliable
structure for compilation. &
Compile-time diagnostic with runtime fallback to the original package context. \\
\bottomrule
\end{tabular}
\end{table}

\name{} uses a hybrid classifier for this step.  It first extracts structural
features from \(d\) and \(\mathcal{A}\), including headings, ordered lists,
command blocks, script manifests, function signatures, asset types, and
verification phrases.  
A compile-time LLM then judges the source shape from
these grounded features and emits a label with supporting evidence.  
The final
decision applies a conservative priority order: strong workflow evidence yields
a workflow graph; independent callable assets yield a dispatcher;
reference-dominant packages yield indexed guidance; and packages with weak or
conflicting evidence yield an insufficient diagnostic.

\vspace{-0.2cm}
\subsection{Boundary Contract as the Public Skill ABI}

The \emph{artifact boundary contract} is the public runtime ABI of a compiled
skill.  It exposes operators, input/output requirements, risk and validation
metadata, and action policies while hiding compiler-local graphs and extraction
details.  Figure~\ref{fig:boundary_contract} summarizes the contract as a
static record consumed by the shared runtime.
All source-shape lowerings are normalized into this contract before the artifact becomes visible to the runtime.

\begin{figure}[h]
\centering
\begin{tikzpicture}[
  node distance=0mm,
  header/.style={
    draw,
    rounded corners=2.5pt,
    align=center,
    text width=0.82\linewidth,
    inner sep=4pt,
    font=\small,
    fill=blue!8
  },
  field/.style={
    draw,
    align=left,
    text width=0.82\linewidth,
    minimum height=6.5mm,
    inner sep=3pt,
    font=\scriptsize
  }
]
\node[header] (title)
  {\textbf{Artifact Boundary Contract} \(B\)};
\node[field, fill=black!2, below=of title] (type)
  {\textbf{type \(\tau\)}: guidance, adapter, typed operator, or solver-like boundary};
\node[field, fill=black!1, below=of type] (ops)
  {\textbf{operators \(O\)}: exposed callable units with schemas, bindings, and source references};
\node[field, fill=black!2, below=of ops] (io)
  {\textbf{I/O \(C_{io}\)}: required arguments, outputs, and task-bound inputs};
\node[field, fill=black!1, below=of io] (risk)
  {\textbf{risk/validation \(R,V\)}: execution risks and evidence level};
\node[field, fill=black!2, below=of risk] (policy)
  {\textbf{policies \(\pi_a,\pi_s\)}: action ranking and operator-selection hints};
\node[field, fill=black!1, below=of policy] (fallback)
  {\textbf{fallback \(F\)}: lossless metadata for returning to the source package};
\end{tikzpicture}
\caption{Boundary contract as a static runtime ABI record.}
\label{fig:boundary_contract}
\end{figure}

Formally, we represent the boundary contract as
\[
B = (\tau, O, C_{io}, R, V, \pi_a, \pi_s, F),
\]
where \(\tau\) is the boundary type, \(O\) is the set of exposed operators,
\(C_{io}\) is the input/output contract, \(R\) is a set of risk flags, \(V\)
is the validation level, \(\pi_a\) is the action policy, \(\pi_s\) is the
selection policy, and \(F\) is lossless fallback metadata.

Together, these fields specify a declarative ABI for compiled skills. 
The ABI
defines the interfaces exposed by the compiled artifact together with the
evidence and policy metadata associated with those interfaces. 
It does not
encode a complete task solution, nor does it require a particular execution
trace. 
Validation is represented as an ordered evidence level, ranging from syntactic conformance checks to executable tests, verifier-backed checks, and regression evidence. 
Thus, validation records the strength of the evidence available for an artifact, rather than serving as a formal correctness guarantee.
Accordingly, our use of policy and validation metadata should be understood as engineering guardrails for guarded execution, not as a formal safety guarantee.

\subsection{Boundary-Aware Runtime Execution}

\begin{figure}
    \centering
    \includegraphics[width=0.75\linewidth]{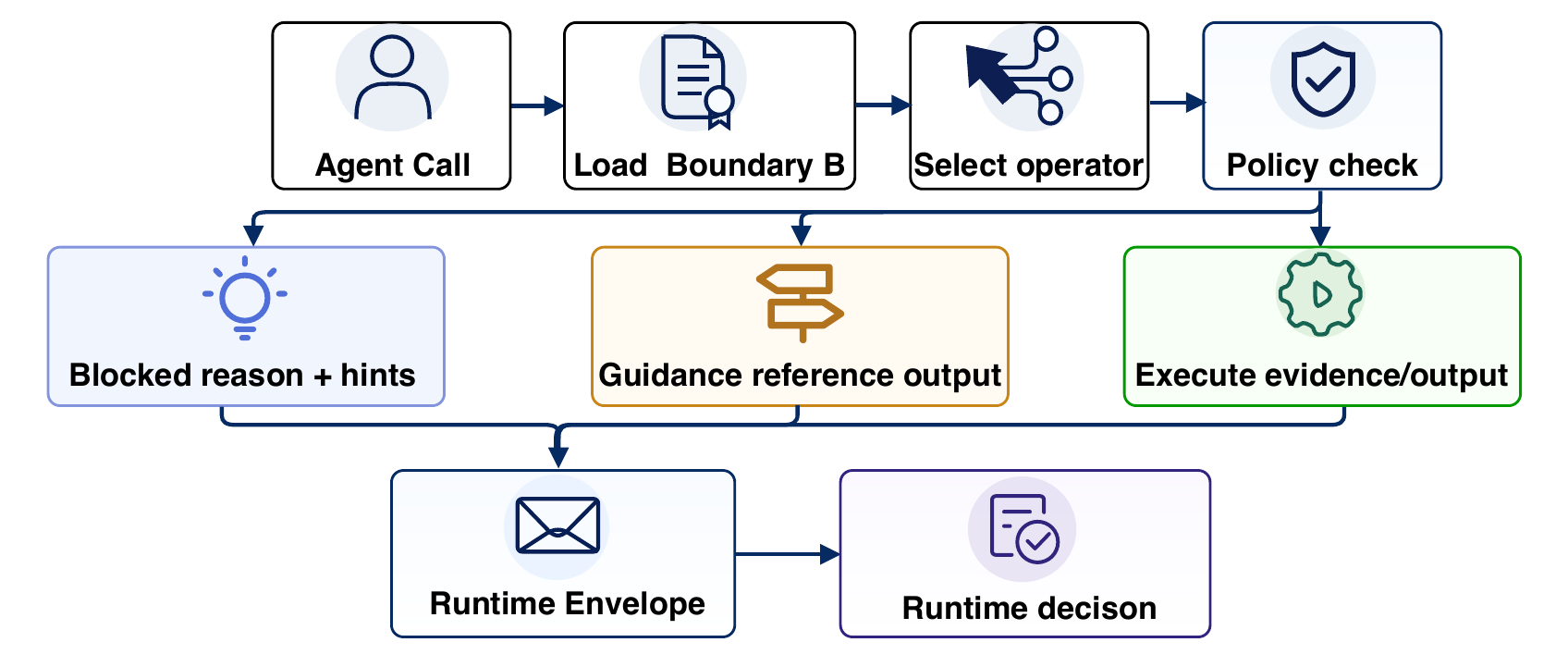}
    \caption{Runtime interpretation of a boundary contract as a guarded state machine.}
    \vspace{-0.2cm}
    \label{fig:runtime_state_machine}
\end{figure}

Given a boundary contract and a runtime invocation, the shared boundary runtime
interprets the contract as a guarded state machine
(Figure~\ref{fig:runtime_state_machine}).  The boundary-contract subsection
defines what a compiled artifact exposes; this subsection defines how the
runtime consumes that interface on each call.  The generated artifact remains
thin: it stores skill metadata and \(B\), while the shared runtime handles selection, policy checks, execution, and envelope construction.

The three policy outcomes differ in commitment.  \texttt{blocked} returns a
reason and deoptimization hints.  
\texttt{guidance} returns reference guidance
and requires agent continuation.  
\texttt{execute} runs a typed operator, script
asset, or command and returns either typed evidence or solver output.  
All paths
return the same canonical envelope, containing status, contribution type,
selected operator, output, trace, and a continuation flag.  Thus, runtime
execution is partial by construction: a compiled artifact can help without
silently claiming that it solved the task.

\subsection{Lossless Deoptimization and Agent Integration}

\name{} treats compilation as a non-destructive transformation.  When lowering
cannot express all source guidance as executable operators, \name{} stores the
original package in a lossless capsule with its assets, text chunks, hashes,
and package identity.  The boundary publishes this capsule through standard
deoptimization operators, including \texttt{list\_skill\_assets},
\texttt{get\_skill\_asset}, and \texttt{search\_skill\_docs}.  The runtime and
agent use these operators to recover exact source material whenever compiled
structure remains incomplete, policy prevents execution, or the current task
requires information outside the compiled operators.

\name{} integrates with an agent through the same progressive-disclosure pattern
used for raw skills.  It advertises each compiled artifact through a compact
\texttt{run\_\{skill\}} handle and boundary summary, while keeping detailed
operator schemas, policy hints, and fallback contents outside the persistent
system prompt.  Once the agent selects a compiled skill, the runtime discloses
the relevant boundary fields on demand and executes the selected operator when
policy permits.  This integration positions \name{} between raw skill retrieval
and full solver synthesis: it extracts reusable structure at compile time,
invokes typed execution when safe, and preserves a lossless return path to the
source package when agent reasoning must continue.

\section{Experiments}
\label{sec:exp}

\subsection{Setup}

\vspace{-0.2cm}
\textbf{Benchmark.}
We evaluate \name{} on SkillsBench~\cite{li2026skillsbench}, a benchmark for
measuring how effectively agents use skill packages. SkillsBench tasks provide
task instructions, task-local skill folders, environment assets, and
deterministic verifiers. In our checkout, the default runnable task set contains
87 tasks and 227 task-local skill packages. This structure matches our goal:
it lets us measure not only final task success, but also how much repeated runtime skill interpretation costs under a fixed tool environment and verifier.

\textbf{Tasks.}
From SkillsBench~\cite{li2026skillsbench}, we select seven verified skill instances as a compact representative evaluation suite. The selection is stratified by task difficulty, covering easy, medium, and hard tasks, and is chosen to reflect the benchmark's typical workload mix: structured document generation, semantic verification, numerical/file manipulation, and long-horizon multimodal or scientific processing.
The hard group contains
\texttt{3d-scan-calc}, \texttt{mars-clouds-clustering}, and
\texttt{video-tutorial-indexer}; these tasks require long-horizon execution
over specialized inputs such as binary geometry files, scientific annotations,
or videos. 
For example, \texttt{mars-clouds-clustering} requires searching
DBSCAN hyperparameters with a custom distance metric and extracting a Pareto
frontier. 
The medium group contains \texttt{citation-check},
\texttt{jax-computing-basics}, and \texttt{pptx-reference-formatting}; these
tasks combine deterministic computation or file manipulation with semantic
judgment. 
For example, \texttt{citation-check} asks the agent to identify fake
or hallucinated papers from plausible BibTeX metadata. 
The easy group contains
\texttt{offer-letter-generator}, a structured document-generation task that
fills a Word template from employee data. We provide detailed descriptions of
all seven tasks in Appendix~\ref{app:task-details}.

\textbf{Agent Harnesses.}
We evaluate three host agent harnesses. 
The primary harness is an internally developed production-grade agent, anonymized as \textsc{Agent-H}, which provides
a fixed tool-calling interface, sandboxed execution environment, and task-local workspace for all methods. 
We also evaluate Codex~\cite{openai2026codexcli},
OpenAI's terminal-based coding agent, and OpenCode~\cite{opencode2026}, an open-source coding agent for terminal, IDE, and desktop workflows. 
Unless otherwise specified, \name{} and all baselines use the same model, agent harness, sandbox, and tool environment in each comparison.

\textbf{Models.}
We evaluate four models spanning frontier, efficient medium-scale, and small
model regimes: GPT-5.5~\cite{openai2026gpt55}, Claude Opus
4.7~\cite{anthropic2026claudeopus47}, DeepSeek V4
Flash~\cite{deepseek2026v4flash}, and Qwen3.6 35B
A3B~\cite{qwen2026qwen3635ba3b}. 
For \name{}, compile-time model cost is
reported separately and included in repeated-invocation amortization experiments.

\textbf{Platform.}
We use a MacBook Pro with 32GB memory running macOS Tahoe 26.3.1 as the local
execution platform, which is representative of common terminal-based agent
workflows. 
We route all model calls through OpenRouter~\cite{openrouter2026}
for all evaluated models, so differences across methods do not come from
changing provider access paths.
We repeat each experimental condition five times and report mean solve-stage metrics over these repeated runs. 
Unless otherwise stated, the token, latency, and iteration values in the main figures and text are averages over the five runs.

\textbf{Baselines.}
We compare \name{} against three baselines.
\textbf{Raw-Skills} exposes the original \texttt{SKILL.md} files and bundled
assets to the agent at runtime, matching the standard interpretation-based
skill-use setting.
\textbf{SkVM-Compiled Skills} uses SkVM~\cite{chen2026skvm}, a skill virtual
machine that compiles skills for heterogeneous LLMs and agent harnesses through
capability profiling, environment binding, and workflow-oriented optimization.
This baseline compares \name{} with a recent compiler-style skill execution
system rather than with prompt-only skill use.
All baselines use the same underlying model and tool environment unless
otherwise specified. For each model tier, we run the full baseline suite so
that improvements cannot be attributed to a stronger model or a different tool
configuration.

\textbf{Metrics.}
We report token usage, number of LLM calls, end-to-end latency, monetary cost, task success rate, execution path consistency, and failure localization accuracy. 
Task success is measured by the SkillsBench verifier for each task. 
Token usage and LLM calls are collected from the agent and model-provider logs. We report monetary cost as a token-metered proxy under paired comparisons that use the same runtime model, provider route, and tool environment; under this setting, monetary cost is monotonic with token usage and is not intended as an independent cross-provider pricing study. For repeated-invocation experiments, we additionally report amortized token-metered cost including the one-time compilation overhead.
To complete the details of our experiments and enhance reproducibility, we have included additional details in the Appendix~\ref{app:reproducibility}.

\vspace{-0.2cm}
\subsection{Overall Runtime Benefits}
\label{sec:overall-benefits}

\begin{figure}[t]
    \centering
    \includegraphics[width=1\linewidth]{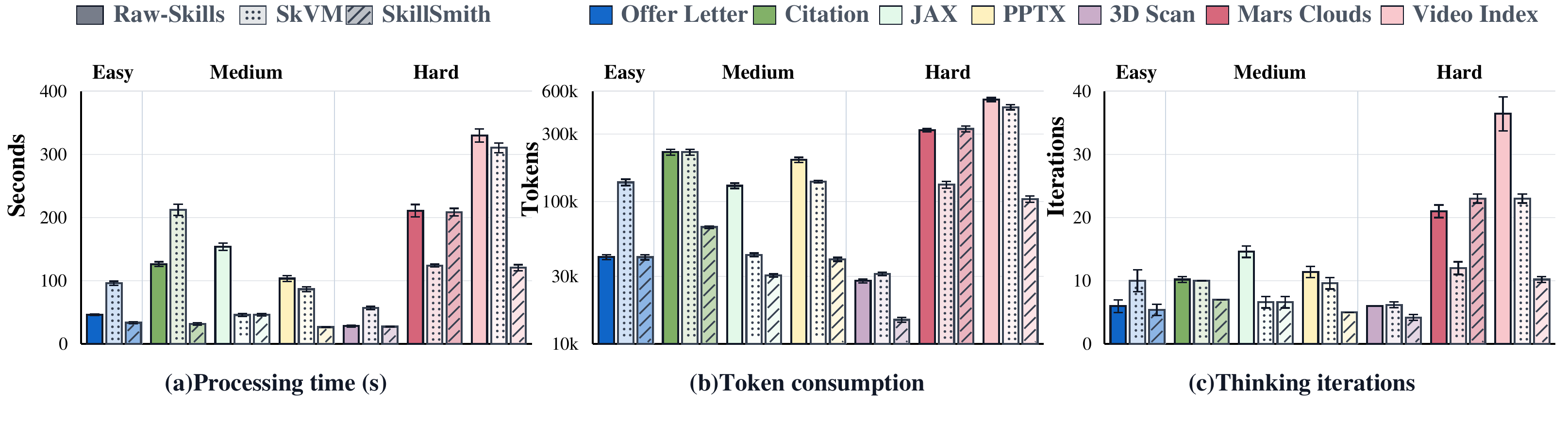}
    \caption{Overall runtime benefits across seven SkillsBench tasks.} 
    \label{fig:overall-benefits}
    \vspace{-0.4cm}
\end{figure}

Figure~\ref{fig:overall-benefits} provides an overall view of the runtime
benefits of \name{} across seven representative SkillsBench tasks.  
All bars report five runs.
In this experiment, all methods use GPT-5.5 as the underlying model.  
The comparison overs wall-clock processing time, total token consumption, and the number of agent thinking iterations, which together capture the cost of repeatedly
interpreting skills at runtime.  
\name{} solves all seven tasks while using 620K total tokens, 494 seconds, and 61 thinking iterations.  
Compared with Raw-Skills, which also solves all tasks but uses 1.5M tokens,
999 seconds, and 107 iterations, \name{} reduces token usage by 57.44\%,
runtime by 50.57\%, and thinking iterations by 42.99\%.
Compared with SkVM, which uses 1.2M tokens, 933 seconds, and
75 iterations, \name{} reduces token usage by 46.49\%, runtime by 47.04\%,
and thinking iterations by 18.67\%.  
The no-skills setting uses 699,181 tokens and 923.556 seconds, but solves only 6 of 7 tasks, 
so we treat it as a runtime context rather than a success-equivalent skill-use baseline.  
These results show that compiling skills into boundary-governed runtime interfaces reduces
the token and deliberation overhead of skill use without sacrificing task success.
Task-level supplementary results for Figure~\ref{fig:overall-benefits} are
reported in Appendix~\ref{app:task-level-results}.

\vspace{-0.2cm}
\subsection{Model Stability}
\vspace{-0.4cm}

\begin{figure}[h]
    \centering
    \includegraphics[width=0.9\linewidth]{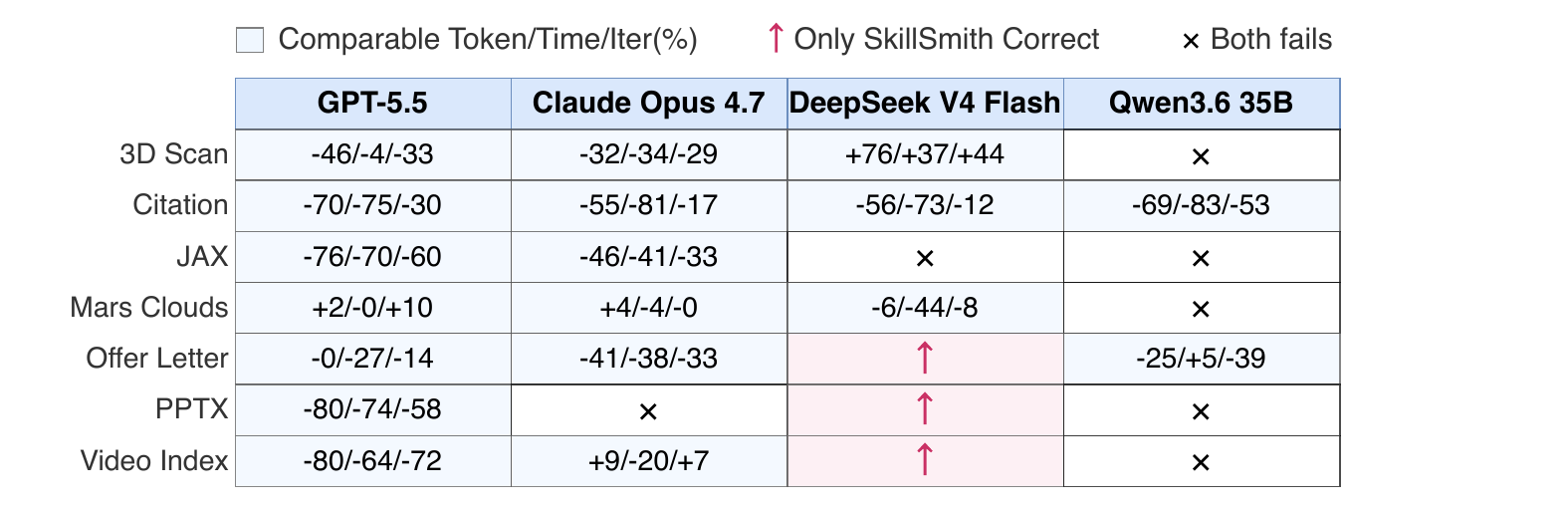}
    \vspace{-0.1cm}
    \caption{Cross-model correctness and runtime benefit summary.}
    \label{fig:cross_model_runtime_benefits}
\end{figure}
We further evaluate whether the benefits of compiled skills are stable across
different underlying models.
All compiled-skill artifacts used in these runs are generated with Claude Opus
4.7 and then reused across the different runtime models.
Figure~\ref{fig:cross_model_runtime_benefits}
reports results on four models: GPT-5.5, Claude Opus 4.7, DeepSeek V4 Flash,
and Qwen3.6 35B. 
For success-preserving comparisons where both Raw Skills and 
\name{} pass the verifier, \name{} maintains Raw-Skills correctness while
reducing execution cost. 
Across the 18 comparable cells in the table, \name{}
achieves an average time reduction of 38.33\%, 
token reduction of 32.83\%, and
iteration reduction of 23.89\%. 
In the table, these savings are shown as
negative signed changes because lower runtime cost is better. 
Beyond preserving
correctness, \name{} can also improve accuracy: on DeepSeek V4 Flash, using
compiled artifacts produced by Claude Opus 4.7 enables \name{} to solve
\texttt{offer-letter-generator}, \texttt{pptx-reference-formatting}, and
\texttt{video-tutorial-indexer}, all of which Raw Skills fails under the same
runtime model. 
This suggests that compilation can transfer skill structure from
a stronger compile-time model to a weaker or more efficient runtime model,
yielding both efficiency and accuracy gains.
Task-level supplementary results for Figure~\ref{fig:cross_model_runtime_benefits}
are reported in Appendix~\ref{app:task-level-results}.

\vspace{-0.2cm}
\subsection{Harness Stability}
\begin{wrapfigure}{r}{0.42\linewidth}
    \vspace{-0.15in}
    \centering
    \resizebox{\linewidth}{!}{%
    \begin{tikzpicture}[
        x=0.62cm,y=0.034cm,
        bar/.style={draw=black, line width=0.25pt},
        tokenbar/.style={bar, fill=tokenblue},
        timebar/.style={bar, fill=timepink},
        iterbar/.style={bar, fill=iteryellow},
        tick/.style={font=\fontsize{7.6}{8.0}\selectfont, text=black!70},
        ylabel/.style={font=\fontsize{9.6}{10.0}\selectfont, text=black!70},
        label/.style={font=\fontsize{10.1}{10.5}\selectfont, text=black!75},
        vallabel/.style={font=\fontsize{5.7}{6.0}\selectfont\bfseries, text=black!78},
        nalabel/.style={font=\fontsize{7.7}{8.0}\selectfont\bfseries, text=black!50},
        legend/.style={font=\fontsize{9.9}{10.4}\selectfont, text=black!75}
    ]

    \draw[black!70, line width=0.45pt] (0.55,0) rectangle (9.95,90);
    \foreach \y in {0,20,40,60,80} {
        \draw[black!10, line width=0.18pt] (0.55,\y) -- (9.95,\y);
        \node[tick, anchor=east] at (0.30,\y) {\y};
    }

    \node[ylabel, rotate=90, anchor=center] at (-0.65,42.5) {Reduction (\%)};

\draw[draw=black!55, line width=0.25pt, fill=tokenblue] (1.1,93.0) rectangle (1.7,99);
\node[legend, anchor=west] at (1.8,95.9) {Tokens};

\draw[draw=black!55, line width=0.25pt, fill=timepink] (4.1,93.0) rectangle (4.7,99);
\node[legend, anchor=west] at (4.8,95.9) {Time};

\draw[draw=black!55, line width=0.25pt, fill=iteryellow] (7,93) rectangle (7.6,99);
\node[legend, anchor=west] at (7.7,95.9) {Iterations};

    \draw[tokenbar] (1.05,0) rectangle (1.705,55.8);
    \draw[timebar]  (1.82,0) rectangle (2.475,27.4);
    \draw[iterbar]  (2.59,0) rectangle (3.245,7.6);
    \node[vallabel] at (1.3775,60.6) {55.8};
    \node[vallabel] at (2.1475,32.2) {27.4};
    \node[vallabel] at (2.9175,12.4) {7.6};
    \node[label, align=center] at (2.1475,-13) {OpenCode};

    \draw[tokenbar] (4.15,0) rectangle (4.805,77.0);
    \draw[timebar]  (4.92,0) rectangle (5.575,52.7);
    \node[vallabel] at (4.4775,81.8) {77.0};
    \node[vallabel] at (5.2475,57.5) {52.7};
    \node[nalabel] at (6.0175,7.0) {N/A};
    \node[label, align=center] at (5.2475,-13) {Codex};

    \draw[tokenbar] (7.25,0) rectangle (7.905,57.4);
    \draw[timebar]  (8.02,0) rectangle (8.675,50.6);
    \draw[iterbar]  (8.79,0) rectangle (9.445,43.0);
    \node[vallabel] at (7.5775,62.2) {57.4};
    \node[vallabel] at (8.3475,55.4) {50.6};
    \node[vallabel] at (9.1175,47.8) {43.0};
    \node[label, align=center] at (8.3475,-13) {\textsc{Agent-H}};

    \end{tikzpicture}}
    \caption{Harness-level solve-stage reductions.}
    \vspace{-0.1cm}
    \label{fig:harness-reductions}
    \vspace{-0.1in}
\end{wrapfigure}
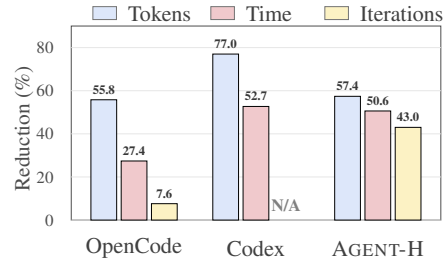

We evaluate whether compiled skills remain effective across agent harnesses in Figure~\ref{fig:harness-reductions}.
Reported reductions are computed from mean solve-stage metrics over five runs.
We run OpenCode, Codex, and \textsc{Agent-H} with GPT-5.5 on the seven verified SkillsBench tasks, comparing each raw-skills harness against the same compiled skills.
Relative to OpenCode Raw, compiled skills reduce solve-stage tokens by 55.8\%, time by 27.4\%, and measured iterations by 7.6\%.
Relative to Codex Raw, they reduce tokens by 77.0\% and time by 52.7\%; Codex iterations are N/A because the CLI trace does not expose internal LLM iteration counts.
Relative to \textsc{Agent-H} Raw, they reduce tokens by 57.4\%, time by 50.6\%, and iterations by 43.0\%.
In general, token reductions exceed end-to-end time reductions because \name{} reduces reasoning and skill-interpretation work, but not task tool-call time; once tool execution is included, wall-clock gains are diluted.
Harness-level supplementary benefit summaries for Figure~\ref{fig:harness-reductions} are
reported in Appendix~\ref{app:task-level-results}.

\vspace{-0.2cm}
\subsection{Compilation Cost}
We measure the one-time compilation cost on 9 different skills.  
Across
the GPT-5.5 and Claude Opus 4.7 compilation runs, compiling one reusable
artifact costs 3,104 tokens and 13.22 seconds on average.  The corresponding
same-model raw-to-compiled runtime comparisons reduce aggregate solve-stage
token use from 742K to 343K tokens, a 53.79\% reduction, and reduce
solve time from 476 to 234 seconds, a 50.73\% reduction.  
This
compilation cost is paid once per compiled skill artifact, while the runtime
savings recur on every subsequent invocation, so the compile overhead can be
amortized across repeated uses of the same skill.

\vspace{-0.2cm}
\subsection{Limitations}
\name{} targets the reusable structure inside skill rather than the entire task execution. Its effectiveness therefore depends on the source skill exposing stable procedures, constraints, and assets that can be
compiled into a useful runtime boundary. When a skill is incomplete, outdated,
or tightly coupled to assumptions that are not present in the deployment
environment, the compiled artifact can inherit those mismatches. Compiled
artifacts are also tied to the tool versions, file formats, dependency
environment, and execution policies under which they were produced, so changes
to those boundaries should trigger revalidation or recompilation. In practice,
\name{} is most helpful when repeated calls spend substantial effort
interpreting and coordinating reusable skill logic; it does not eliminate
irreducible work such as external tool execution, file I/O, media processing, or task-specific judgment that must still happen at runtime.
We also discussed broader impacts in the Appendix~\ref{app:broader}.

\vspace{-0.2cm}
\section{Conclusion}

We presented \name{}, a boundary-first compiler-runtime framework that lowers
reusable agent skills from repeatedly interpreted \texttt{SKILL.md} context into
minimal runtime interfaces exposing only the operators, inputs, policies,
validation evidence, and fallback paths needed for guarded, policy-governed use.
On seven SkillsBench tasks, \name{} reduces solve-stage token usage by 57.44\%,
thinking iterations by 42.99\%, solve time by 50.57\% (2.02$\times$ faster), and token-proportional monetary cost by 57.44\% compared with Raw-Skills; 
it also reduces tokens by 46.49\%, solve time by 47.04\%, and iterations by 18.67\% compared with the SkVM. Across evaluated harnesses and models, \name{} preserves
Raw-Skills correctness and improves success tasks on medium-scale models by reusing artifacts compiled with a stronger model. 
These results suggest that
compiling skills into boundary-governed runtime interfaces can improve both the efficiency and portability of skill-using LLM agents.

\bibliographystyle{plain}
\bibliography{references}

\appendix

\newpage
\section{Additional Method Details}
\label{app:method-details}

\subsection{Lowering Decisions}
\label{app:lowering}

The lowering stage uses structural signals in the skill package to choose among workflow, dispatcher, and reference artifacts. Workflow evidence includes enumerated procedures, imperative verbs, command blocks, explicit dependencies, and sections such as ``steps'', ``workflow'', or ``procedure''. Dispatcher evidence includes a collection of scripts, APIs, command templates, or independent callable utilities with no dominant execution order. Reference evidence includes high prose ratio, style guidance, domain background, examples, and weak executable structure.

The output is a conservative lowering decision rather than only a class label. Workflow skills are compiled into typed graphs. Dispatcher skills are lowered into callable registries with routing metadata. Reference skills are indexed as retrieval-oriented guidance. This design avoids forcing all skills into a single representation.

\subsection{Workflow IR and Validation}
\label{app:workflow-ir}

Each workflow node contains a step identifier, human-readable name, step type, execution specification, input and output contracts, dependency list, provenance metadata, risk annotation, and cacheability flag. Source references link a node back to relevant spans in \texttt{SKILL.md}, command blocks, scripts, or packaged assets. These references are used both for debugging and for failure localization.

After LLM-assisted completion, \name{} applies normalization and validation. Normalization canonicalizes step identifiers, removes duplicated nodes, repairs simple dependency omissions, and standardizes step-type names. Validation checks that every edge points to an existing node, the graph is acyclic, executable nodes contain runnable specifications, and required inputs are produced by earlier nodes or provided by the invocation. Nodes that cannot be grounded or validated are either repaired conservatively or marked for runtime model assistance rather than executed deterministically.

\subsection{Runtime Guarantees}
\label{app:runtime-guarantees}

\name{} provides four practical runtime guarantees. First, deterministic steps are executed from compiled specifications rather than regenerated online. Second, source-grounded provenance allows failures to be traced to the responsible step, command, and skill span. Third, isolated sandboxes package local dependencies and reduce interference between skills. Fourth, graph execution enables partial recovery and reuse: after a failure, the runtime can resume from the failed node when upstream cached outputs remain valid.

These guarantees are intentionally systems-level guarantees. \name{} does not claim formal correctness of arbitrary generated workflows. Instead, it narrows runtime uncertainty by making execution structure explicit, validating the graph before execution, and limiting open-ended LLM calls to nodes whose semantics genuinely require generation.

\section{Additional Experiment Details}
\label{app:exp-details}

\subsection{Task Details}
\label{app:task-details}

We provide detailed descriptions of the seven SkillsBench tasks used in our
evaluation. The tasks are grouped by difficulty.

\paragraph{Task-local skills.}
Each evaluated task exposes one or more task-local skill packages under its
\texttt{environment/skills/} directory. These packages are the skill inputs used
by Raw-Skills, SkVM-Compiled Skills, and \name{}. We list all skills used in the
seven-task evaluation below.

\begin{itemize}
    \item \textbf{\texttt{3d-scan-calc}} uses
    \textbf{\texttt{mesh-analysis}}, which provides a Python
    \texttt{MeshAnalyzer} for binary STL parsing, connected-component analysis,
    largest-component extraction, volume computation, and material-attribute
    extraction from STL attribute bytes. The skill is used to filter scan debris
    and compute the geometric quantities needed for mass estimation.

    \item \textbf{\texttt{mars-clouds-clustering}} uses three skills.
    \textbf{\texttt{custom-distance-metrics}} explains how to define callable
    application-specific distances for clustering algorithms such as
    \texttt{sklearn.cluster.DBSCAN}, including parameterized distance functions
    and distance-matrix construction. \textbf{\texttt{parallel-processing}}
    provides \texttt{joblib}-based patterns for parallel grid search and batch
    computation across CPU cores. \textbf{\texttt{pareto-optimization}}
    describes Pareto dominance, Pareto-frontier extraction, use of the
    \texttt{paretoset} library, manual non-dominated filtering, and
    visualization of trade-off solutions.

    \item \textbf{\texttt{video-tutorial-indexer}} uses
    \textbf{\texttt{speech-to-text}}, which provides a timestamped
    transcription workflow for videos using a pre-installed Whisper tiny model.
    The skill exposes a script interface that converts the tutorial video into
    timestamped transcript segments for downstream chapter alignment.

    \item \textbf{\texttt{citation-check}} uses
    \textbf{\texttt{citation-management}}, a citation verification and metadata
    management skill. It covers paper search, DOI/PMID/arXiv metadata
    extraction, BibTeX generation, duplicate detection, reference cleaning, and
    validation of citation fields such as title, authors, venue, year, and DOI.

    \item \textbf{\texttt{jax-computing-basics}} uses
    \textbf{\texttt{jax-skills}}, which provides JAX-oriented numerical
    computing guidance and helper APIs for loading and saving arrays, map and
    reduce operations, logistic-loss gradients, recurrent scans, JIT execution,
    shape validation, and numerically stable array processing.

    \item \textbf{\texttt{pptx-reference-formatting}} uses
    \textbf{\texttt{pptx}}, a PowerPoint creation, editing, and analysis skill.
    It covers text extraction, raw Office Open XML access, unpacking and packing
    \texttt{.pptx} archives, slide XML editing, validation, theme and typography
    inspection, and preservation of presentation layout while modifying slide
    content.

    \item \textbf{\texttt{offer-letter-generator}} uses
    \textbf{\texttt{docx}}, a Word document manipulation skill based on
    \texttt{python-docx}. It focuses on robust placeholder replacement when
    placeholders are split across XML runs, processing paragraphs, tables,
    nested tables, headers, footers, and conditional document sections.
\end{itemize}

\paragraph{Hard tasks.}
These tasks require long-horizon execution over specialized inputs and involve
multiple tightly coupled processing steps.

\begin{itemize}
    \item \textbf{\texttt{3d-scan-calc}} asks the agent to compute the mass of a
    scanned 3D-printed part from a binary STL file. The main challenges are
    binary STL parsing, connected-component filtering to remove scanning debris,
    volume calculation, material-density lookup, and unit consistency.

    \item \textbf{\texttt{mars-clouds-clustering}} asks the agent to optimize
    DBSCAN hyperparameters for Mars cloud annotation clustering. The main
    challenges are implementing a custom distance metric, performing a large
    grid search, matching citizen-science clusters to expert labels, aggregating
    metrics, and extracting the Pareto frontier.

    \item \textbf{\texttt{video-tutorial-indexer}} asks the agent to identify
    chapter start times in a 23-minute Blender tutorial video. The main
    challenges are multimodal video understanding, timestamp alignment, and
    preserving the exact order of 29 fixed chapter titles.
\end{itemize}

\paragraph{Medium tasks.}
These tasks combine deterministic computation or file manipulation with
semantic judgment.

\begin{itemize}
    \item \textbf{\texttt{citation-check}} asks the agent to detect fake or
    hallucinated papers in a BibTeX file. The main challenge is verifying
    plausible citation metadata using titles, authors, venues, years, DOIs, and
    external evidence.

    \item \textbf{\texttt{jax-computing-basics}} asks the agent to solve
    numerical programming problems using JAX and save the required
    \texttt{.npy} outputs. The main challenge is mapping natural-language
    computation requests to correct JAX operations, array shapes, and dtypes.

    \item \textbf{\texttt{pptx-reference-formatting}} asks the agent to reformat
    dangling paper titles in a PowerPoint deck and add a reference slide. The
    main challenge is editing PowerPoint XML while preserving layout,
    formatting, and deduplicating paper references.
\end{itemize}

\paragraph{Easy task.}
This task mainly tests structured document generation.

\begin{itemize}
    \item \textbf{\texttt{offer-letter-generator}} asks the agent to fill a Word
    offer-letter template from employee data. The main challenge is replacing
    placeholders without breaking document structure and correctly handling a
    conditional relocation section.
\end{itemize}

\subsection{Task-Level Supplementary Results}
\label{app:task-level-results}

The following tables provide task-level values underlying
Figures~\ref{fig:overall-benefits} and~\ref{fig:cross_model_runtime_benefits},
plus harness-level task benefit summaries for Figure~\ref{fig:harness-reductions}.
Each compact result cell in the task-level tables reports
\emph{correctness; tokens/time/iterations}. Correctness is denoted by P for a passed verifier and F for a failed verifier.

\begin{table}[p]
\centering
\scriptsize
\caption{Task-level values corresponding to Figure~\ref{fig:overall-benefits}.
Each cell reports correctness; solve tokens / solve time in seconds / LLM
iterations.}
\label{tab:fig5-task-level}
\resizebox{\linewidth}{!}{%
\begin{tabular}{lcccc}
\toprule
\textbf{Task} & \textbf{No Skills} & \textbf{Raw-Skills} & \textbf{\name{}} & \textbf{SkVM} \\
\midrule
3D Scan & F; 32K/35.4/5 & P; 28K/28.6/6 & P; 15K/27.4/4 & P; 31K/57.1/6 \\
Citation & P; 98K/199.9/8 & P; 223K/126.6/10 & P; 66K/32.0/7 & P; 222K/211.8/10 \\
JAX & P; 60K/57.4/11 & P; 129K/153.4/15 & P; 30K/46.0/6 & P; 43K/46.1/6 \\
Mars Clouds & P; 52K/80.7/6 & P; 318K/209.1/21 & P; 324K/208.9/23 & P; 128K/123.8/12 \\
Offer Letter & P; 53K/97.6/10 & P; 41K/46.1/7 & P; 41K/33.6/6 & P; 137K/96.3/9 \\
PPTX & P; 243K/234.8/16 & P; 197K/103.7/12 & P; 39K/26.5/5 & P; 139K/87.4/9 \\
Video Index & P; 162K/217.7/20 & P; 524K/332.5/36 & P; 105K/119.9/10 & P; 461K/310.9/23 \\
\bottomrule
\end{tabular}}
\end{table}

\begin{table}[p]
\centering
\scriptsize
\caption{Task-level cross-model values corresponding to
Figure~\ref{fig:cross_model_runtime_benefits}. Each result cell reports
correctness; solve tokens / solve time in seconds / LLM iterations. The flag
column records task-level metric differences and Raw-Skills failures fixed by
\name{}.}
\label{tab:fig6-task-level}
\resizebox{\linewidth}{!}{%
\begin{tabular}{llccl}
\toprule
\textbf{Runtime model} & \textbf{Task} & \textbf{Raw-Skills} & \textbf{\name{}} & \textbf{Flag} \\
\midrule
GPT-5.5 & 3D Scan & P; 28K/28.6/6 & P; 15K/27.4/4 & -- \\
GPT-5.5 & Citation & P; 223K/126.6/10 & P; 66K/32.0/7 & -- \\
GPT-5.5 & JAX & P; 129K/153.4/15 & P; 30K/46.0/6 & -- \\
GPT-5.5 & Mars Clouds & P; 318K/209.1/21 & P; 324K/208.9/23 & tok, iter \\
GPT-5.5 & Offer Letter & P; 41K/46.1/7 & P; 41K/33.6/6 & -- \\
GPT-5.5 & PPTX & P; 197K/103.7/12 & P; 39K/26.5/5 & -- \\
GPT-5.5 & Video Index & P; 524K/332.5/36 & P; 105K/119.9/10 & -- \\
Claude & 3D Scan & P; 61K/22.3/7 & P; 42K/14.7/5 & -- \\
Claude & Citation & P; 196K/99.7/6 & P; 89K/19.4/5 & -- \\
Claude & JAX & P; 104K/39.7/9 & P; 57K/23.5/6 & -- \\
Claude & Mars Clouds & P; 338K/104.2/15 & P; 349K/100.5/15 & tok \\
Claude & Offer Letter & P; 230K/61.6/15 & P; 135K/38.5/10 & -- \\
Claude & PPTX & P; 1280K/166.2/30 & F; 69K/21.2/5 & regression \\
Claude & Video Index & P; 219K/143.8/14 & P; 239K/114.8/15 & tok, iter \\
DeepSeek & 3D Scan & P; 68K/43.5/9 & P; 120K/59.4/13 & tok, time, iter \\
DeepSeek & Citation & P; 530K/287.5/17 & P; 232K/77.1/15 & -- \\
DeepSeek & JAX & P; 218K/101.5/16 & F; 223K/89.6/19 & regression \\
DeepSeek & Mars Clouds & P; 549K/559.1/24 & P; 517K/313.5/22 & -- \\
DeepSeek & Offer Letter & F; 86K/64.1/10 & P; 392K/108.3/21 & fixes raw \\
DeepSeek & PPTX & F; 2973K/351.7/60 & P; 984K/236.1/35 & fixes raw \\
DeepSeek & Video Index & F; --/--/-- & P; 366K/264.4/23 & fixes raw \\
Qwen & 3D Scan & P; 522K/150.0/27 & F; 184K/140.9/17 & regression \\
Qwen & Citation & P; 349K/295.5/17 & P; 109K/51.8/8 & -- \\
Qwen & JAX & P; 304K/111.2/25 & F; 153K/210.8/15 & regression \\
Qwen & Mars Clouds & F; 131K/57.3/13 & F; 277K/464.8/15 & -- \\
Qwen & Offer Letter & P; 175K/69.1/18 & P; 131K/72.3/11 & time \\
Qwen & PPTX & F; 209K/101.3/12 & F; 241K/234.9/14 & -- \\
Qwen & Video Index & F; 227K/297.5/14 & F; 75K/462.2/8 & -- \\
\bottomrule
\end{tabular}}
\end{table}

\begin{table}[p]
\centering
\scriptsize
\caption{Task-level harness reductions corresponding to
Figure~\ref{fig:harness-reductions}. Values are percentage reductions relative
to the \textsc{Agent-H} \name{} reference and do not report raw solve-stage
measurements. Codex iteration reduction is N/A because the CLI trace does not
expose internal LLM iteration counts.}
\label{tab:fig7-harness-benefit}
\resizebox{\linewidth}{!}{%
\begin{tabular}{lccccccccc}
\toprule
& \multicolumn{3}{c}{\textbf{Tokens}} &
\multicolumn{3}{c}{\textbf{Time}} &
\multicolumn{3}{c}{\textbf{Iterations}} \\
\cmidrule(lr){2-4}\cmidrule(lr){5-7}\cmidrule(lr){8-10}
\textbf{Task} & \textbf{OpenCode} & \textbf{Codex} & \textbf{\textsc{Agent-H}} &
\textbf{OpenCode} & \textbf{Codex} & \textbf{\textsc{Agent-H}} &
\textbf{OpenCode} & \textbf{Codex} & \textbf{\textsc{Agent-H}} \\
\midrule
3D Scan & 86.0\% & 91.8\% & 46.4\% & 18.5\% & 57.2\% & 4.2\% & 50.0\% & N/A & 33.3\% \\
Citation & 76.7\% & 79.6\% & 70.4\% & 28.9\% & 62.0\% & 74.7\% & 12.5\% & N/A & 30.0\% \\
JAX & 78.4\% & 87.7\% & 76.7\% & 16.7\% & 51.5\% & 70.0\% & 33.3\% & N/A & 60.0\% \\
Mars Clouds & 5.8\% & 45.8\% & -1.9\% & 27.2\% & 33.8\% & 0.1\% & -53.3\% & N/A & -9.5\% \\
Offer Letter & 48.1\% & 80.5\% & 0.0\% & 8.2\% & 63.1\% & 27.1\% & 0.0\% & N/A & 14.3\% \\
PPTX & 85.2\% & 92.3\% & 80.2\% & 67.7\% & 84.5\% & 74.4\% & 54.5\% & N/A & 58.3\% \\
Video Index & 44.7\% & 83.4\% & 80.0\% & 15.0\% & 46.4\% & 63.9\% & -11.1\% & N/A & 72.2\% \\
\bottomrule
\end{tabular}}
\end{table}

\clearpage

\subsection{Additional Reproducibility Information}
\label{app:reproducibility}

We provide additional details for reproducing the experimental results reported
in Section~4. The public source code and data are released at the anonymous
repository listed in the main paper. A reproduction run should use the same
SkillsBench checkout, the same seven task identifiers, the same task-local skill
folders and assets, and the deterministic SkillsBench verifiers used in our
evaluation.

\paragraph{Benchmark subset and task inputs.}
All main experiments use the following seven SkillsBench tasks:
\texttt{3d-scan-calc}, \texttt{mars-clouds-clustering},
\texttt{video-tutorial-indexer}, \texttt{citation-check},
\texttt{jax-computing-basics}, \texttt{pptx-reference-formatting}, and
\texttt{offer-letter-generator}. The SkillsBench checkout used in our
experiments contains 87 runnable tasks and 227 task-local skill packages. We do
not alter task instructions, bundled assets, verifier code, or expected output
formats. Each trial is run in a fresh task-local workspace so that intermediate
files, cached outputs, and failed attempts from one trial cannot affect another
trial. For \name{}, compiled skill artifacts are produced from the task-local
\texttt{SKILL.md} files, bundled assets, tool specification, and execution
environment fingerprint described in Section~3; the raw task inputs and
verifiers remain unchanged.

\paragraph{Methods compared.}
The primary comparison includes Raw-Skills, SkVM-Compiled Skills, and
\name{}. Raw-Skills exposes the original \texttt{SKILL.md} files and bundled
skill assets to the agent at runtime. SkVM-Compiled Skills uses the SkVM
compiler-style skill execution baseline under the same runtime model and tool
environment. \name{} first compiles each skill package into a boundary-governed
runtime interface and then exposes the compiled interface during solving. The
no-skills condition is used only as a runtime context in the overall comparison,
because it solves only six of the seven selected tasks.

\paragraph{Models, harnesses, and provider route.}
The main overall experiment uses GPT-5.5 as the runtime model. Cross-model
experiments evaluate GPT-5.5, Claude Opus 4.7, DeepSeek V4 Flash, and Qwen3.6
35B A3B. Unless a comparison explicitly studies cross-model artifact reuse, all
methods in a paired comparison use the same runtime model, harness, sandbox,
tool environment, and provider route. For the cross-model reuse experiment, the
compiled artifacts are generated once with Claude Opus 4.7 and reused with each
runtime model. All model calls are routed through OpenRouter. We evaluate three
agent harnesses: the anonymized \textsc{Agent-H} harness, Codex, and OpenCode.
Codex iteration counts are not reported because its CLI trace does not expose
internal LLM iteration counts.

\paragraph{Execution platform.}
All local executions are run on a MacBook Pro with 32GB memory running macOS
Tahoe 26.3.1. The platform is used for sandboxed command execution, file I/O,
document processing, numerical scripts, media processing, and verifier
execution. Because wall-clock time includes both model latency and local tool
execution, exact latency can vary with provider load and local machine state.
For this reason, we report averages over repeated trials and compare methods
under paired settings that share the same provider route and local execution
environment.

\paragraph{Trial protocol.}
Each experimental condition is repeated five times. A condition is defined by a
task, method, runtime model, harness, and compiled-artifact setting. For each
trial, we start from the original task workspace, run the agent until completion
or failure under the same harness policy, and then invoke the task verifier.
For \name{}, compilation cost is recorded separately from solve-stage cost. In
solve-stage figures, we report runtime solving metrics only; in amortization
experiments, we additionally include the one-time compilation overhead. Across
the GPT-5.5 and Claude Opus 4.7 compilation runs, compiling one reusable
artifact costs 3,104 tokens and 13.22 seconds on average.

\paragraph{Metrics and aggregation.}
Task success is determined by the deterministic SkillsBench verifier for the
corresponding task. Token usage and LLM-call counts are collected from agent and
provider logs. Runtime is measured as end-to-end solve-stage wall-clock time,
including model calls, tool calls, local script execution, file reads/writes,
and verifier-facing artifact generation. Thinking iterations are counted from
the harness trace when the harness exposes this quantity; otherwise the value is
reported as unavailable. Monetary cost is reported as a token-metered proxy
within paired comparisons that use the same runtime model, provider route, and
tool environment. We aggregate token usage, time, and iteration counts by
averaging the five trials for each condition, and the main figures report these
mean values unless stated otherwise.

\paragraph{Verifier and failure handling.}
A run is counted as successful only when the corresponding SkillsBench verifier
accepts the produced output. If an agent terminates without producing the
required artifact, produces an artifact with the wrong format, or produces an
artifact rejected by the verifier, the run is counted as a failure. Runtime
metrics are still retained for failed runs so that success-rate and cost results
reflect the complete experimental condition. For traceability and failure
localization analyses, \name{} records the compiled step, command or operator,
and source provenance span associated with each runtime action; these records
are compared against the observed failing step when a run fails.

\paragraph{Reproduction checklist.}
To reproduce the reported numbers, use the released repository and run the seven
task identifiers listed above with five repeated trials per condition. Keep the
SkillsBench task files and verifiers unchanged; use the same model route
through OpenRouter; run Raw-Skills, SkVM-Compiled Skills, and \name{} under the
same harness and sandbox for each paired comparison; separate \name{}'s
compile-time metrics from solve-stage metrics unless reproducing amortized
cost; and aggregate token, time, and iteration values as means over the five
trials. Reproductions should expect small wall-clock variation from provider
latency and local machine load, while verifier outcomes and relative token
accounting should remain stable under the same model and harness configuration.

\section{Broader Impacts}
\label{app:broader}

\name{} can reduce the computational and monetary cost of repeated skill-based
agent execution, which may lower the environmental and financial overhead of
LLM-agent workflows and make reproducible automation more accessible. Its
explicit boundary contracts and source-provenance records can also improve
debugging, auditability, and failure localization for deployed agents. More
generally, compiling reusable skill structure into explicit runtime interfaces
can make agent behavior easier to inspect than repeated free-form interpretation
of long skill documents.

At the same time, compiled skills can amplify mistakes when source skills
contain incorrect assumptions, stale tool dependencies, unsafe procedures, or
environment-specific behavior that no longer holds at deployment time. A
compiled artifact may make an erroneous workflow faster and cheaper to repeat,
and lower-cost automation can also be misused to scale harmful or
policy-violating agent workflows. There is also a risk of over-reliance:
operators may treat compiled artifacts as correctness guarantees even though
\name{} only validates structural and execution boundaries, not the semantic
truth of every downstream action.

We therefore view compiled artifacts as execution aids rather than standalone
guarantees of safety or correctness. Practical mitigations include sandboxed
execution, verifier-based validation, provenance inspection, explicit artifact
boundary contracts, and recompilation or revalidation whenever task inputs,
tools, policies, or deployment environments change. In high-stakes settings,
compiled skills should be paired with human review, domain-specific safety
checks, and deployment policies that restrict unsafe or unauthorized workflows.


\end{document}